\documentclass[letterpaper]{article} % DO NOT CHANGE THIS
\usepackage{aaai25}  % DO NOT CHANGE THIS
\usepackage{times}  % DO NOT CHANGE THIS
\usepackage{helvet}  % DO NOT CHANGE THIS
\usepackage{courier}  % DO NOT CHANGE THIS
\usepackage[hyphens]{url}  % DO NOT CHANGE THIS
\usepackage{graphicx} % DO NOT CHANGE THIS
\usepackage{natbib}  % DO NOT CHANGE THIS AND DO NOT ADD ANY OPTIONS TO IT
\usepackage{caption} % DO NOT CHANGE THIS AND DO NOT ADD ANY OPTIONS TO IT
\usepackage{algorithm}
\usepackage{algorithmic}
\usepackage{xcolor}
\usepackage{amsmath}
\usepackage{amssymb}
\usepackage{amsfonts}
\usepackage{multirow}
\usepackage{array}
\usepackage{makecell}
\usepackage{arydshln}
\usepackage{newfloat}
\usepackage{listings}
\DeclareCaptionStyle{ruled}{labelfont=normalfont,labelsep=colon,strut=off} % DO NOT CHANGE THIS
\floatstyle{ruled}
\newfloat{listing}{tb}{lst}{}
\floatname{listing}{Listing}

\usepackage[colorlinks=true, linkcolor=black, citecolor=black, urlcolor=black]{hyperref}

\title{Continual Learning Using a Kernel-Based Method Over Foundation Models}

\author{
    Saleh Momeni\textsuperscript{\rm 1},~
    Sahisnu Mazumder\textsuperscript{\rm 2},~
    Bing Liu\textsuperscript{\rm 1}\\
}
\affiliations{
    \textsuperscript{\rm 1}~Department of Computer Science, University of Illinois Chicago, USA\\
    \textsuperscript{\rm 2}~Intel Labs, USA\\
    smomen3@uic.edu, sahisnumazumder@gmail.com, liub@uic.edu \\
}

\begin{document}

\maketitle

\begin{abstract}
Continual learning (CL) learns a sequence of tasks incrementally. This paper studies the challenging CL setting of \textit{class-incremental learning} (CIL). CIL has two key challenges: \textit{catastrophic forgetting} (CF) and \textit{inter-task class separation} (ICS). Despite numerous proposed methods, these issues remain persistent obstacles.
This paper proposes a novel CIL method, called \textit{Kernel Linear Discriminant Analysis} (KLDA), that can effectively avoid CF and ICS problems. It leverages only the powerful features learned in a \textit{foundation model} (FM). However, directly using these features proves suboptimal. To address this, KLDA incorporates the Radial Basis Function (RBF) kernel and its Random Fourier Features (RFF) to enhance the feature representations from the FM, leading to improved performance.
When a new task arrives, KLDA computes only the mean for each class in the task and updates a shared covariance matrix for all learned classes based on the kernelized features. Classification is performed using \textit{Linear Discriminant Analysis}.
Our empirical evaluation using text and image classification datasets demonstrates that KLDA significantly outperforms baselines. Remarkably, without relying on replay data, KLDA achieves accuracy comparable to joint training of all classes, which is considered the upper bound for CIL performance. The KLDA code is available at \url{https://github.com/salehmomeni/klda}.
\end{abstract}

\section{Introduction}
\label{sec:intro}

Continual learning (CL) is a machine learning paradigm that incrementally learns a sequence of tasks, enabling models to adapt to new data while retaining knowledge from previously learned tasks \cite{chen2018lifelong,de2021continual}. This paper specifically addresses \textit{class-incremental learning} (CIL), a setting where each task introduces a distinct set of classes, and the objective is to train a single unified model capable of recognizing all classes encountered across tasks. A key feature of CIL is that no task identification information is provided during testing, meaning the model must distinguish between classes from different tasks without being informed which task a test sample belongs to.

There are also two other main CL settings: \textit{task-incremental learning} (TIL), which has the task information provided at test time, and \textit{domain-incremental learning} (DIL), which learns tasks from different domains with the same set of classes. For more details about TIL and DIL, please refer to \cite{van2019three}. 

CIL has two major challenges: (1) \textit{catastrophic forgetting} (CF), which occurs when learning new tasks causes the model's performance on earlier tasks to deteriorate due to parameter updates \cite{McCloskey1989}; (2) \textit{inter-task class separation} (ICS) \cite{kim2022theoretical}, which arises because, without access to previous task data while learning a new task, it is difficult to learn decision boundaries between the new and old classes. Many existing CIL methods have been proposed \cite{ke2022continualsurvey,wang2024comprehensive}. 
However, due to the two challenges, they still have significant performance deterioration as more tasks are learned. 

This paper proposes a novel technique using foundation models (FMs) that addresses both challenges and can achieve an accuracy level comparable to the upper bound of CIL -- the accuracy obtained by joint training on all classes/tasks together -- \textit{without} using any replay data.

The proposed method is called \textit{Kernel Linear Discriminant Analysis} (KLDA). In KLDA, the FM is \textbf{frozen}, meaning no updates are made to its parameters to avoid CF. KLDA does not use trainable adapters \cite{houlsby2019parameter} or prompts \cite{wang2022learning} either, as they are also susceptible to CF or ICS. It uses only the latent features of the input samples extracted from the FM. However, directly using these features is suboptimal, as the classes in the original feature space may not be easily separable. To address this, we employ kernel functions, which enhance feature representation by implicitly mapping the data to a higher-dimensional space where the features become more linearly separable. As computing the traditional kernel matrix is impractical for CIL due to its huge size and incremental learning nature, we approximate it using \textit{Random Fourier Features} (RFF) \cite{rahimi2007random}, making it feasible for CIL. When a new task arrives, KLDA uses the \textit{kernalized} features to compute the class \textit{mean} of each class and updates a \textit{shared covariance matrix} for all classes learned thus far. The class mean and the shared covariance matrix define a Gaussian distribution for each class, which \textit{Linear Discriminant Analysis} (LDA) \cite{izenman2013linear} then employs to optimize class separability.

In summary, this paper makes three key contributions:
\begin{enumerate}
\item It introduces a novel method called KLDA which utilizes the rich feature representations from an FM and enhances them using a kernel-based approach. During the CIL process, KLDA incrementally accumulates statistics of the kernelized features and constructs decision boundaries for the classes using LDA. To our knowledge, this methodology has not been previously reported.

\item KLDA overcomes the CF and ICS problems. As KLDA only collects statistics during CIL without updating parameters of the FM or training any additional network, it has no CF. 
Since the classification is based on the feature mean of each class and the shared covariance matrix, the ICS problem disappears because the covariance matrix and the mean define a Gaussian distribution, which naturally separates each class from the rest.

\item Experimental results demonstrate that KLDA significantly outperforms baselines. Most notably, it achieves the accuracy level of fine-tuning the FM on all classes jointly as a single task -- widely considered the upper bound for CIL -- without relying on replay data. This achievement is particularly significant, as existing CIL methods have yet to reach this upper bound, often falling short by a considerable margin. Overcoming this gap is critical, as accuracy remains a paramount factor for the practical adoption of CIL.
\end{enumerate}

\section{Related Work}
\label{sec:related works}

% Continual learning (CL) has emerged as a major research topic of machine learning, with many studies addressing the challenge of catastrophic forgetting (CF). 
Existing CL methods generally fall into several categories:

\textit{Regularization-based methods,} which use regularizers to protect important parameters from major changes when learning new tasks, thereby minimizing CF \cite{kirkpatrick2017overcoming,Zenke2017continual}.

\textit{Replay-based methods,} which store some samples from previous tasks and allow the model to be trained on both new and stored data to maintain performance across tasks \cite{aljundi2019gradient,liu2021lifelong,qin2022elle,huang2021continual}. Some methods use data generators instead of storing actual data, creating samples similar to those from previous tasks \cite{shin2017continual,he2018overcoming}.

\textit{Architectural-based methods,} which involve structural changes in the network. A key approach is \textit{parameter isolation}, where sub-networks are trained for each task, using mechanisms like masking or ensuring orthogonality between parameters of different tasks \cite{serra2018overcoming,gururangan2022demix,geng2021continual,lin2022beyond,wortsman2020supermasks}. Some methods also expand the network as new tasks are introduced \cite{wang2022beef,yan2021dynamically}. Yet, 
some CIL methods employ a task predictor to identify the appropriate model for the predicted task classifier \cite{kim2023learnability,lin2023class,abati2020conditional,wang2023rehearsal}. These systems may utilize strategies like separate networks, entropy, or out-of-distribution detection to determine the task.

The rise of FMs has led to a growing interest in using pre-trained models in CL \cite{yang2024recent}. In NLP, employing pre-trained language models (LM) is a standard approach as they help improve performance significantly \cite{shin2017continual,shao2023class,ke2021achieving}. In computer vision, pre-trained models are also increasingly used \cite{kim2023learnability,lin2023class,wang2022learning}. However, many existing methods used FMs trained using supervised data, which have already covered the classes used in CL, causing serious information leaks \cite{wang2022learning,mcdonnell2024ranpac}. Many also use these FMs within the framework of the CL strategies discussed above, which still suffer from CF and ICS issues. Our approach is different. We explore the full potential of \textit{self-supervised foundation models} as fixed feature extractors. %, i.e., leveraging only their feature extraction capabilities for downstream tasks. 
However, as our experiments reveal, directly using the latent features of the FM is suboptimal for CIL. We then propose to enhance these features through non-linear transformations using kernels, which improves class separability in the kernelized feature space. % The proposed approach, KLDA, is distinct in that it is replay-free and does not rely on regularizers or architectural modifications. 
Our method is related to the streaming method in \cite{hayes2020lifelong}, which uses \textit{streaming linear discriminant analysis} for online CL. However, our KLDA differs by employing a kernel method, and we do not focus on streaming data.

\section{Background}
% This section presents the main background information for the proposed method. 
\subsection{\textbf{Class-incremental Learning}}
In CIL, a model is trained on a sequence of tasks \( \{ \mathcal{T}_1, \mathcal{T}_2, \dots, \mathcal{T}_T \} \), where each task \( \mathcal{T}_t \) introduces a disjoint set of classes with its associated training data \( \mathcal{D}_t = \{(x_t^{(i)}, y_t^{(i)})\}_{i=1}^{N_t} \). The goal is to develop a unified model \( F : \mathcal{X} \rightarrow \mathcal{Y} \) capable of classifying instances from any of the classes encountered across the $T$ tasks.

Because in learning a new task \( \mathcal{T}_t \), the data from previous tasks \( \mathcal{T}_{1}, \dots, \mathcal{T}_{t-1} \) is not accessible, CIL faces two key challenges CF and ICS as mentioned earlier.  
% Thus, the model must learn new classes without forgetting previously learned ones, despite the absence of earlier training data. 
During inference, the task identity is unknown, and the model must predict the correct class label from all the classes encountered so far.

\subsection{Class-prototypes for Continual Learning}

Fine-tuning a foundation model for CIL often leads to \textit{catastrophic forgetting}. Instead, leveraging the latent features to incrementally accumulate \textit{class-prototypes} (CPs) while keeping the FM frozen can result in more accurate classification, as we will demonstrate in the experiments section. A simple yet effective method is the \textit{Nearest Class Mean} (NCM) classifier, where the prototype for a class is the \textit{mean} of the feature vectors extracted from the FM for all training samples of the class. For simplicity, from this point on, we will use $\mathbf{x}$ to denote the feature vector extracted from the FM for an input. The class prototype $\mu_m$ for a class $m$ is then computed as:
\begin{equation}
\mathbf{\mu}_m = \frac{1}{n_m} \sum_{i=1}^{n_m} \mathbf{x}_i
\end{equation}
where $n_m$ is the number of samples for class $m$. This mean vector can be computed incrementally for each class and does not cause CF as it doesn't involve training.

During inference, a test sample is classified by finding the class mean with the highest cosine similarity:
\begin{equation}
\hat{y} = \arg\max_m \frac{\mathbf{x}_{\text{test}}^\top ~\mathbf{\mu}_m}{\|\mathbf{x}_{\text{test}}\| \|\mathbf{\mu}_m\|}
\end{equation}
This straightforward method surprisingly outperforms several more complex prompt-based or fine-tuning-based CIL baselines \cite{zhou2024continual}, which are susceptible to CF. This suggests that FMs provide robust, generalizable representations suitable for downstream tasks.

To enhance NCM, higher-order statistics can be incorporated. A well-known approach is \textit{Quadratic Discriminant Analysis} (QDA) \cite{hastie2009elements}, which assumes that the features of class $m$ follow a \textit{multivariate Gaussian distribution} \( \mathcal{N}(\mathbf{\mu}_m, \mathbf{\Sigma}_m) \). The likelihood \(P(\mathbf{x} \mid y=m)\) can be computed from the Gaussian distribution, and the predicted class is determined by Bayes' rule as the one that maximizes the posterior probability \(P(y \mid \mathbf{x})\). However, this method is unsuitable for continual learning because it requires storing the covariance matrix \(\mathbf{\Sigma}_m\) for every class, which becomes prohibitively large as more classes are introduced in CIL.

\textit{Linear Discriminant Analysis} (LDA) simplifies this by assuming that all classes share the same covariance matrix \( \mathbf{\Sigma} \). %leading to a more tractable model. 
LDA is well-suited for CIL, as it requires storing only the mean vector for each class and a single shared covariance matrix, ensuring that the number of parameters does not grow significantly as the number of classes increases.
The \textit{shared covariance matrix} is computed as follows:
\begin{equation}
\mathbf{\Sigma} = \frac{1}{N} \sum_{m=1}^{M} \sum_{i=1}^{n_m} (\mathbf{x}_{m,i} - \mathbf{\mu}_m)(\mathbf{x}_{m,i} - \mathbf{\mu}_m)^\top
\end{equation}
where $M$ is the number of classes seen so far, and \( N \) is the total number of samples, i.e., \( N = \sum_{m=1}^{M} n_m \). This shared covariance matrix can be updated incrementally in the CIL process when a new task arrives, by first computing the mean for each class before updating \( \mathbf{\Sigma} \). 

Under the shared covariance assumption, the log-posterior for LDA can be written as:
\begin{equation}
\log P(y = m \mid \mathbf{x}) = \mathbf{x}^\top \mathbf{\Sigma}^{-1} \mathbf{\mu}_m - \frac{1}{2} \mathbf{\mu}_m^\top \mathbf{\Sigma}^{-1} \mathbf{\mu}_m + \text{constant}
\end{equation}
Here, the ``constant'' term refers to \( P(\mathbf{x}) \) in Bayes' rule, which is the same for all classes. This approach improves class separation by accounting for the distribution of the data in a class, addressing some limitations of the NCM method.

From this equation, we can define the weight vector \( \mathbf{w}_m \) and bias term \( b_m \) for each class \( m \) as follows:
\begin{equation}
\label{eq.w}
\mathbf{w}_m = \mathbf{\Sigma}^{-1} \mathbf{\mu}_m
\end{equation}
\begin{equation}
\label{eq.b}
b_m = -\frac{1}{2} \mathbf{\mu}_m^\top \mathbf{\Sigma}^{-1} \mathbf{\mu}_m
\end{equation}
This shows that LDA assumes a linear decision boundary, which may not be optimal for features from the FM.

\section{Proposed Method: KLDA}

We propose to enhance LDA with a kernel function, leading to our method, KLDA. The core idea is to improve the linear separability of features from the FM using a kernel function. This allows us to maintain a robust representation of the data as new classes/tasks are added, without suffering from CF.

\subsection{Kernel Functions and Non-linear Transformations}
Linear models often struggle when data is not linearly separable in its original feature space. A powerful approach to overcome this limitation is to use kernel functions, which implicitly map the input data (features from the FM in our case) into a higher-dimensional space where the data becomes more linearly separable. In this high-dimensional space, the model can learn a linear decision boundary that corresponds to a non-linear boundary in the original space.

Mathematically, if we have an input space \( \mathcal{X} \) and a mapping \( \varphi: \mathcal{X} \rightarrow \mathcal{V} \), where \( \mathcal{V} \) is a potentially infinite-dimensional feature space, a kernel function \( K(\mathbf{x}_i, \mathbf{x}_j) \) computes the inner product in this space without explicitly performing the transformation:
\begin{equation} %\[
K(\mathbf{x}_i, \mathbf{x}_j) = \langle \varphi(\mathbf{x}_i), \varphi(\mathbf{x}_j) \rangle_\mathcal{V}
\end{equation} % \]
One of the most commonly used kernels is the Radial Basis Function (RBF), which is defined as:
\begin{equation} %\[
K(\mathbf{x}_i, \mathbf{x}_j) = \exp\left(-\frac{\|\mathbf{x}_i - \mathbf{x}_j\|^2}{2\sigma^2}\right)
\end{equation} % \]
The RBF kernel corresponds to an inner product in an infinite-dimensional space, making it highly effective for capturing complex patterns in data.\footnote{We also experimented with several other kernel functions and found the RBF kernel to be better suited for our CIL setup.} %However, directly computing the kernel matrix \( \mathbf{K} \) for all pairs of instances in a dataset of size \( N \) leads to a large matrix of size \( N \times N \), which is computationally prohibitive. In the continual learning setting, this approach is infeasible as it requires access to data from all previous tasks, which is not available. 
Using this kernel trick involves computing the kernel matrix \( \mathbf{K} \), where \( \mathbf{K}_{ij} = K(\mathbf{x}_i, \mathbf{x}_j) \). This results in an \( N \times N \) matrix, which is computationally prohibitive and impractical in CL. Instead, we approximate the mapping \( \varphi(\mathbf{x}) \) using \textit{Random Fourier Features} and work directly with these finite-dimensional features to avoid computing or storing \( \mathbf{K} \).

\subsection{Approximating the Kernel with Random Fourier Features}
Random Fourier Features \cite{rahimi2007random} provides an efficient way to approximate the kernel function by leveraging Bochner's theorem \cite{rudin2017fourier}, which states that any continuous, shift-invariant kernel can be represented as the Fourier transform of a non-negative measure:
\begin{equation}
    K(\mathbf{x}_i, \mathbf{x}_j) = \int p(\omega) e^{i \omega^\top (\mathbf{x}_i - \mathbf{x}_j)} d\omega = \mathbb{E}_{\omega} \left[ e^{i \omega^\top (\mathbf{x}_i-\mathbf{x}_j)}\right]
\end{equation}
Here, $\omega$ is the frequency in the Fourier domain, and $p(\omega)$ is the probability density function associated with $\omega$. Given that both the kernel $K(\mathbf{x}_i, \mathbf{x}_j)$ and the distribution $p(\omega)$ are real, the integral can be simplified. The complex exponential $e^{i \omega^\top (\mathbf{x}_i-\mathbf{x}_j)}$ can be expressed in terms of its real part using Euler's formula. Therefore, we can obtain a real-valued mapping that satisfies the condition $\mathbb{E}[z_{\omega}(\mathbf{x_i}) z_{\omega}(\mathbf{x_j})] = K(\mathbf{x_i}, \mathbf{x_j})$ by setting:
\begin{equation}
    z_{\omega}(\mathbf{x}) = \sqrt{2} \cos(\omega^\top \mathbf{x} + \beta)
\end{equation}
where $\omega \sim p(\omega)$,  $\beta \sim \text{Uniform}(0, 2\pi)$. For the RBF kernel, the Fourier transform \( p(\omega) \) is a Gaussian distribution \cite{rahimi2007random}. We now have a simple and efficient algorithm to estimate the kerneled features by pooling $D$ independent pairs $\omega$, $\beta$ from these distributions and estimating the expectation. Therefore, we can define the random feature map as:
\begin{equation}
    \mathbf{z}(\mathbf{x}) = \sqrt{\frac{2}{D}} \left[\cos(\omega_1^\top \mathbf{x} + \beta_1), \dots, \cos(\omega_D^\top \mathbf{x} + \beta_D)\right]
\end{equation}
where $\omega$ is drawn from \( \mathcal{N}(0, \sigma^{-2}\mathbf{I}) \) and $\beta$ from \( \text{Uniform}(0, 2\pi) \). As the number of pooled pairs $D$ increases, the approximation of the kernel function improves because more Monte Carlo samples are used to estimate the expectation. The dot product of these random features approximates the original kernel function:
\begin{equation}
    \mathbf{z}(\mathbf{x}_i)^\top \mathbf{z}(\mathbf{x}_j) \approx K(\mathbf{x}_i, \mathbf{x}_j)
\end{equation}
Thus, \( \mathbf{z} \) represents an approximation of \( \varphi \). We can now convert the input \( \mathbf{x} \) into random features \( \mathbf{z}(\mathbf{x}) \) and apply linear methods. This approximation enables us to avoid directly computing the kernel matrix, making it feasible to apply in continual learning settings while preserving the benefits of the kernel transformation.

\subsection{Classification with KLDA}
\textbf{Training:} The training process of KLDA is outlined in Algorithm~\ref{alg:OKLDA}. We first apply RFF to the original feature vector \(\mathbf{x} \in \mathbb{R}^d\), transforming it into \(\mathbf{z} \in \mathbb{R}^D\). With each new class, the mean \(\mathbf{\mu}_m\) is calculated, and the shared covariance matrix \(\mathbf{\Sigma}\) is updated incrementally.

\vspace{1mm}
\noindent \textbf{Prediction:} We compute the linear coefficients \( \mathbf{w}_m \) (Eq.~\ref{eq.w}) and bias \( b_m \) (Eq.~\ref{eq.b}) based on the mean vectors and the shared covariance. These parameters are aggregated into a weight matrix \( \mathbf{W} = [\mathbf{w}_m]_{m=1}^M \) and a bias vector \( \mathbf{b} = [b_m]_{m=1}^M \). The classification function in the transformed space is:
\begin{equation} %\[
F(\mathbf{x}) = \mathbf{z}(\mathbf{x})^\top \mathbf{W} + \mathbf{b}
\end{equation} %\]
Here, \( F(\mathbf{x}) \) provides the score for each class, and the predicted class is the one with the highest score value.

We also introduce an \textbf{ensemble} approach, \textbf{KLDA-E}, to further enhance performance. It leverages multiple KLDA models, each initialized with distinct frequency matrices and phase vectors (Algorithm~\ref{alg:OKLDA}). During inference, the scores are calculated for each model and then transformed into probabilities via \textbf{softmax}. The final prediction is made by averaging the probabilities of all models and selecting the class with the highest average probability: %\footnote{We also experimented with other ways of combining predictions from multiple models, such as hard voting, but found this approach to be the most effective, as it resolves tie votes.}
\begin{equation}
\hat{y} = \arg\max_m \frac{1}{E} \sum_{e=1}^{E} P_e(y = m \mid \mathbf{x})
\end{equation}

% By leveraging RFF for kernel approximation and applying LDA in the transformed space, KLDA ensures robust class separation and learns new classes incrementally, while avoiding CF.

\begin{algorithm}[t]
\caption{KLDA Training}
\label{alg:OKLDA}
\begin{algorithmic}[1] %[1] enables line numbers
\STATE \textbf{Initialize}
\STATE $\omega \sim \mathcal{N}(0, {\sigma^{-2}\mathbf{I}}) \in \mathbb{R}^{d \times D}$ \hfill \COMMENT{RFF frequency matrix}
\STATE $\beta \sim \mathcal\mathbb{U}(0, 2\pi) \in \mathbb{R}^{D}$ \hfill \COMMENT{RFF phase vector}
\STATE $\mathbf{\Sigma} = \mathbf{0} \in \mathbb{R}^{D \times D}$ \hfill \COMMENT{shared covariance matrix}
\STATE $N_{\text{total}} = 0$ \hfill \COMMENT{total number of samples}

\vspace{2mm}
\STATE \textbf{Function} RFF($X$):
\STATE \hspace{1em} \textbf{return} $\sqrt{\frac{2}{D}} \cos(X \omega + \beta)$ \hfill \COMMENT{RFF applied to batch}

\vspace{0.15cm}
\STATE \textbf{Function} Update($X$, $m$):
\STATE \hspace{1em} \textbf{Input:} $\mathbf{X} \in \mathbb{R}^{n_m \times d}$ - batch feature vectors for all training samples of class $m$
\STATE \hspace{1em} $N_{\text{prev}} \gets N_{\text{total}}$
\STATE \hspace{1em} $N_{\text{total}} \gets N_{\text{total}} + n_m$
\STATE \hspace{1em} $Z \gets \text{RFF}(X)$
\STATE \hspace{1em} Compute class mean: $\mathbf{\mu}_m \gets \frac{1}{n_m} \sum_{i=1}^{n_m} Z_i$
\STATE \hspace{1em} Update covariance matrix: 
\STATE \hspace{2em} $\mathbf{\Sigma} \gets \frac{N_{\text{prev}}}{N_{\text{total}}} \mathbf{\Sigma} + \frac{1}{N_{\text{total}}} \sum_{i=1}^{n_m} (Z_i - \mathbf{\mu}_m)(Z_i - \mathbf{\mu}_m)^\top$
\end{algorithmic}
\end{algorithm}

\noindent \textbf{Theoretical Justification:} It has been shown theoretically that good within-task prediction (WP) and effective out-of-distribution (OOD) detection for the tasks learned so far are \textit{necessary} and \textit{sufficient} conditions for good CIL \cite{kim2022theoretical,kim2023learnability}. In our approach, since each class is represented as a Gaussian distribution, effectively, each task has only one class. Then, WP is always correct and the Gaussian distribution serves as an OOD detector for each class. %. CIL depends only on OOD detection of each class.

\section{Experimental Setup}
This section outlines the datasets, baselines, implementation details, and evaluation metrics employed in our experiments.

\subsection{Datasets}
We conduct experiments on both \textbf{text} and \textbf{image} classification datasets to evaluate our proposed method. However, our primary focus is text classification, as language foundation models (LFMs) are more mature. For our main experiments, we use the following four text classification datasets:

\begin{itemize}
    \item \textbf{CLINC}: It has 150 classes of dialogue intents from many different application domains \cite{larson2019evaluation}. We used the train/test split of 10,000/750 samples, and the classes were randomly divided into 10 disjoint tasks.
    \item \textbf{Banking}: It has 77 classes of dialogue intents in the banking domain \cite{casanueva2020efficient}. We used a 10,000/1,000 train/test split and divided the classes into 7 disjoint tasks. 
    \item \textbf{DBpedia}: A text classification dataset of Wikipedia articles with 70 classes \cite{liu2021benchmarking}. We used a train/test split of 10,000/1,000 samples and divided the classes into 7 disjoint tasks.
    \item \textbf{HWU}: Another dialogue intent classification dataset featuring 20 domains with 64 classes \cite{auer2007dbpedia}. We used a train/test split of 9,000/1,000 samples and partitioned the classes into 8 disjoint tasks. 
\end{itemize}

We follow the typical CIL protocol. The classes in each dataset are randomly shuffled and assigned to the tasks. We perform multiple runs with random shuffles to account for the variability due to different task splits.

\textbf{Image Datasets}: We also evaluate KLDA using four image classification datasets: \textbf{CIFAR10} and \textbf{CIFAR100} with 10 and 100 classes respectively \cite{krizhevsky2009learning}, \textbf{TinyImageNet} with 200 classes \cite{le2015tiny}, and Stanford \textbf{Cars} with 196 classes \cite{yang2015large}, applying their official train/test splits. The comparison is made against the CIL upper bound, which jointly trains on all tasks simultaneously. Since KLDA supports the incremental addition of CPs, task splits are not required for this evaluation.

\subsection{Baselines}
For text classification, we compare our approach against multiple baselines, categorized into fine-tuning based methods, class-prototype methods, and joint training.

\subsubsection{Fine-tuning Based Baselines}
\begin{itemize}
    \item \textbf{Vanilla}: It sequentially fine-tunes the model on each task with no mechanism to mitigate forgetting.
    
    \item \textbf{EWC}: A  regularization-based method that uses a penalty to preserve important parameters from previous tasks to mitigate forgetting \cite{kirkpatrick2017overcoming}.
    
    \item \textbf{KD}: It uses knowledge distillation to help the model retain information from old tasks by learning from softened output probabilities of previous versions of itself \cite{hinton2015distilling}.
    
    \item \textbf{L2P}: Freezes the model and learns trainable prompts to guide inference, adapting to new tasks without altering the base model \cite{wang2022learning}.
    
    \item \textbf{LAMOL}: Employs pseudo-replay by generating pseudo-examples of previous tasks to mix with new task data, maintaining past performance while learning new tasks \cite{sun2019lamol}.
    
    \item \textbf{VAG}: It leverages vocabulary sparsity to mask the probability of unused tokens when training on a task, mitigating forgetting via label generation rather than the traditional classification objective. \cite{shao2023class}.
\end{itemize}

\subsubsection{Class-prototype Based Baselines}

\begin{itemize}
    \item \textbf{NCM}: It maintains a mean feature vector for each class, added incrementally. Classification is based on the distance to the mean vectors.
    
    \item \textbf{LDA}: This utilizes the original feature space without our RBF kernel extension.
\end{itemize}

\subsubsection{Joint Training Baseline}
\begin{itemize}
    \item \textbf{Joint Fine-tuning}: Fine-tuning the FM by updating its parameters and a classifier head added on top of the latent features, training on all classes simultaneously as a single task. This approach is regarded as the \textbf{upper-bound performance} of CIL.
\end{itemize}

% These diverse baselines allow us to thoroughly evaluate the performance of KLDA. 

\subsection{Implementation Details}

For the main experiments, we use BART-base \cite{lewis2019bart}, which consists of a 6-layer encoder-decoder architecture with a 768-dimensional hidden state. This LFM was chosen because most of our fine-tuning baselines use a generative objective or require generating pseudo-replay data during training. Additionally, the same FM is also used in the state-of-the-art VAG system \cite{shao2023class}.

To demonstrate the versatility of KLDA, we also evaluate it with multiple other LFMs, including paraphrase-MiniLM-L3 \cite{reimers2019sentence} (3 layers, 384 dimensions), BERT-base \cite{kenton2019bert} (12 layers, 768 dimensions), RoBERTa-large \cite{liu2019roberta} (24 layers, 1024 dimensions), T5-3b \cite{raffel2020exploring} (24 layers, 1024 dimensions), and Mistral-7b \cite{jiang2023mistral} (32 layers, 4096 dimensions).

For vision foundation models (VLMs), we use the DINOv2-small (12 layers, 384 dimensions) and DINOv2-base (12 layers, 768 dimensions) models \cite{oquabdinov2}. We chose DINOv2 because it is a self-supervised model, unlike commonly used ViT \cite{dosovitskiy2020image} and ResNet \cite{he2016deep} models, which are trained in a supervised manner on ImageNet-21k or ImageNet-1k \cite{deng2009imagenet}. This supervised training can lead to \textbf{information leakage} due to class overlap with datasets used in CIL.

LAMOL and VAG were executed using their official codes and configurations. For the remaining fine-tuning baselines, we used implementations from \cite{shao2023class} repository. The class-prototype-based baselines were implemented using our own code, adhering to the same update rules applied in KLDA to ensure consistency in comparison.

The \textbf{Joint Fine-tuning} method, representing the upper bound, is trained for 50 epochs with a batch size of 128, using the Adam optimizer with a learning rate of 1e-3 for the classifier head and 1e-4 for the FM parameters. Additionally, we experimented with various configurations, including different learning rates, batch sizes, and epoch numbers, to ensure the models were \textit{thoroughly trained and optimized}.

For our ensemble approach KLDA-E, we use a set of 5 models. KLDA has two hyperparameters itself: the transformation dimension \( D \) and the RFF \( \sigma \). Given the CIL setup, where tasks are learned incrementally, the system does not see all tasks at the same time, and validation sets are not typically available. Therefore, it is hard to optimize the parameters for all tasks. Through empirical testing, we found that setting \( D \) to 5000 offers a balanced trade-off between memory usage and performance. The \( \sigma \) parameter is also empirically determined within range \([10^{-2}, 10^{-6}]\) for each FM. We will show the results of different parameter settings later.

Our implementation is built using PyTorch, with all pre-trained models sourced from the Hugging Face Transformers library. All experiments were conducted on a single NVIDIA A100 GPU with 80GB of VRAM.

\subsection{Evaluation Metric}
We measure classification accuracy after all tasks have been processed, referred to as \textbf{Last or Final Accuracy} like \cite{shao2023class}. Each experiment is repeated three times with different random seeds, and the average accuracy is reported to ensure reliable results. We also discuss the efficiency and the memory requirement of the proposed method.

\begin{table*}[ht]
\begin{center}
\scalebox{0.9}{ % fontsize 9
\begin{tabular}{l|cccc}
\hline
\textbf{Method} & \textbf{CLINC} (10-T) & \textbf{Banking} (7-T) & \textbf{DBpedia} (7-T) & \textbf{HWU} (8-T) \\
\hline
% Joint Linear Probe & 95.33 $\pm$ 0.04 & 91.50 $\pm$ 0.24 & 94.73 $\pm$ 0.16 & 88.43 $\pm$ 0.12 \\
Joint Fine-tuning & 95.33 $\pm$\scriptsize{0.04} & 91.36 $\pm$\scriptsize{0.32} & 94.83 $\pm$\scriptsize{0.16} & 88.60 $\pm$\scriptsize{0.29} \\
\hdashline
Vanilla & 42.06 $\pm$\scriptsize{1.53} & 31.80 $\pm$\scriptsize{1.20} & 43.45 $\pm$\scriptsize{2.54} & 30.95 $\pm$\scriptsize{3.37} \\
EWC & 45.73 $\pm$\scriptsize{0.46} & 38.40 $\pm$\scriptsize{2.70} & 44.99 $\pm$\scriptsize{2.90} & 34.01 $\pm$\scriptsize{3.46} \\
KD & 36.33 $\pm$\scriptsize{0.86} & 27.40 $\pm$\scriptsize{1.59} & 42.10 $\pm$\scriptsize{2.40} & 25.46 $\pm$\scriptsize{2.13} \\
L2P & 30.66 $\pm$\scriptsize{2.46} & 31.45 $\pm$\scriptsize{0.55} & 23.52 $\pm$\scriptsize{1.54} & 24.04 $\pm$\scriptsize{0.88} \\
LAMOL & 58.42 $\pm$\scriptsize{0.84} & 42.60 $\pm$\scriptsize{1.36} & 48.61 $\pm$\scriptsize{1.82} & 44.85 $\pm$\scriptsize{1.57} \\
VAG & 76.42 $\pm$\scriptsize{0.90} & 59.34 $\pm$\scriptsize{1.28} & 65.40 $\pm$\scriptsize{1.52} & 56.88 $\pm$\scriptsize{1.22} \\
\hdashline
NCM & 83.60 $\pm$\scriptsize{0.00} & 71.10 $\pm$\scriptsize{0.00} & 75.70 $\pm$\scriptsize{0.00} & 73.30 $\pm$\scriptsize{0.00} \\
LDA & 93.71 $\pm$\scriptsize{0.00} & 89.09 $\pm$\scriptsize{0.00} & 93.42 $\pm$\scriptsize{0.00} & 86.41 $\pm$\scriptsize{0.00} \\
\hline
KLDA & 95.90 $\pm$\scriptsize{0.68} & 92.23 $\pm$\scriptsize{0.32} & 94.13 $\pm$\scriptsize{0.32} & 87.27 $\pm$\scriptsize{1.39} \\
KLDA--E & \textbf{96.62 $\pm$\scriptsize{0.08}} & \textbf{93.03 $\pm$\scriptsize{0.06}} & \textbf{94.53 $\pm$\scriptsize{0.12}} & \textbf{89.78 $\pm$\scriptsize{0.09}} \\
\hline
\end{tabular}
}
\caption{Final accuracy (\%) of different methods on text classification datasets. All results are with a BART-base model, and no replay buffer was used for any method. The number of tasks is indicated in parentheses next to each dataset (\#-T). Note that the number of tasks does not affect NCM, LDA, or KLDA as these methods add a class prototype at a time. Joint Fine-tuning is considered the upper bound for CIL performance since it learns all classes together as a single task.}
\label{tab:main-results}
\end{center}
\vspace{-2mm}
\end{table*}

\section{Experiment Results}
This section evaluates KLDA in terms of accuracy, memory usage, and efficiency across multiple text classification baselines. Additionally, KLDA's performance is compared to the Joint upper bound on image classification datasets, along with an analysis of the impact of hyperparameters on its performance.

\subsection{Comparison with Baselines}
Table \ref{tab:main-results} compares the performance of KLDA with various baselines, including regularization techniques (EWC, KD), prompt-based methods (L2P), and pseudo-replay methods (LAMOL, VAG). Despite specialized mechanisms for mitigating CF, these methods still exhibit significant forgetting, with even the best-performing method, VAG, falling short of the accuracy achieved by the simple NCM method.

NCM, while effective, significantly underperforms compared to Joint, indicating that merely accumulating a mean feature vector for each class is insufficient to fully leverage the information in the FM representations. LDA improves class separation by incorporating a shared covariance and KLDA improves this by leveraging the kernel. The addition of the ensemble approach further enhances accuracy.

KLDA-E consistently matches the accuracy of the Joint Fine-tuning upper bound, even surpassing it on 3 out of the 4 datasets, and achieving nearly identical results on the fourth (DBpedia). Notably, even KLDA alone performs on par with Joint Fine-tuning. This shows that the features of FMs are well-suited for accurate CIL, and the key lies in how to utilize these features appropriately, which is achieved by the proposed method.

\subsection{Memory Usage Comparison}
We compare the methods in terms of memory usage. The Joint Fine-tuning only adds a classifier head on top of the FM features, introducing approximately 0.1M additional parameters for typical values of $M = 150$ classes and $d = 768$ hidden dimensions. Fine-tuning baselines, particularly those required to operate in the generation mode, significantly increase memory usage. For instance, an LM head required for text generation adds approximately 38.5M parameters for a vocabulary of 50,265 tokens, although this number does not increase with the number of classes.

CP methods are more memory-efficient as they only require storing the class prototypes. NCM requires $M \times d$ parameters for the mean vectors, similar to the classifier head of the Joint model. LDA adds an $d \times d$ covariance matrix, increasing the parameter count by approximately 0.6M. KLDA introduces $D \times (d + 1)$ fixed parameters for the RFF transformation. With $D$ set to 5000, this adds around 3.8M parameters. KLDA also scales the parameters required for CPs by a factor of $D/d$, leading to an additional 0.75M parameters. The covariance matrix for KLDA is $D \times D$, resulting in an additional 25M parameters. In total, KLDA’s memory footprint is approximately 29.5M parameters. This memory requirement is still significantly lower than the LM head needed for text generation alone. Our ensemble method utilizes 5 models, resulting in a 5x increase in memory usage, which remains within a reasonable limit. For reference, the BART-base used in our main experiments has 139.5M parameters. We highlight that a large portion of KLDA's parameters are associated with the fixed RFF transformation and the shared covariance matrix, which do not increase as more classes are added to the CIL process.

\begin{table}[h]
\begin{center}
% \resizebox{\columnwidth}{!}{
\scalebox{0.9}{ % fontsize 9
\begin{tabular}{>{\centering\arraybackslash}p{2.4cm}|>{\centering\arraybackslash}p{1.9cm}|>{\centering\arraybackslash}p{1.5cm}|>{\centering\arraybackslash}p{1.5cm}}
\hline
\textbf{Model} & \textbf{Dataset} & \textbf{KLDA-E} & \textbf{Joint} \\
\hline
\multirow{4}{*}{\makecell[c]{\textbf{MiniLM} \\ \scriptsize{3 layers} \\ \scriptsize{384 dimensions}}} & CLINC & 94.53$\pm$\scriptsize{0.00} & 93.20$\pm$\scriptsize{0.16} \\
 & Banking & 91.73$\pm$\scriptsize{0.09} & 90.90$\pm$\scriptsize{0.08} \\
 & DBpedia & 86.83$\pm$\scriptsize{0.17} & 87.43$\pm$\scriptsize{0.16} \\
 & HWU & 87.95$\pm$\scriptsize{0.23} & 87.13$\pm$\scriptsize{0.12} \\
\hline
\multirow{4}{*}{\makecell[c]{\textbf{BERT-base} \\ \scriptsize{12 layers} \\ \scriptsize{768 dimensions}}} & CLINC & 94.98$\pm$\scriptsize{0.31} & 94.56$\pm$\scriptsize{0.04} \\
 & Banking & 91.00$\pm$\scriptsize{0.24} & 88.96$\pm$\scriptsize{0.16} \\
 & DBpedia & 95.40$\pm$\scriptsize{0.08} & 95.03$\pm$\scriptsize{0.09} \\
 & HWU & 88.32$\pm$\scriptsize{0.31} & 87.26$\pm$\scriptsize{0.28} \\
\hline
\multirow{4}{*}{\makecell[c]{\textbf{RoBERTa-large} \\ \scriptsize{24 layers} \\ \scriptsize{1024 dimensions}}} & CLINC & 96.31$\pm$\scriptsize{0.06} & 95.96$\pm$\scriptsize{0.30} \\
 & Banking & 92.93$\pm$\scriptsize{0.05} & 91.16$\pm$\scriptsize{0.04} \\
 & DBpedia & 94.60$\pm$\scriptsize{0.08} & 94.99$\pm$\scriptsize{0.21} \\
 & HWU & 89.25$\pm$\scriptsize{0.04} & 88.40$\pm$\scriptsize{0.29} \\
\hline
\multirow{4}{*}{\makecell[c]{\textbf{T5-3b} \\ \scriptsize{24 layers} \\ \scriptsize{1024 dimensions}}} & CLINC & 96.04$\pm$\scriptsize{0.17} & 96.86$\pm$\scriptsize{0.06} \\
 & Banking & 93.77$\pm$\scriptsize{0.05} & 92.30$\pm$\scriptsize{0.10} \\
 & DBpedia & 95.33$\pm$\scriptsize{0.09} & 94.60$\pm$\scriptsize{0.03} \\
 & HWU & 89.31$\pm$\scriptsize{0.27} & 90.30$\pm$\scriptsize{0.10} \\
\hline
\multirow{4}{*}{\makecell[c]{\textbf{Mistral-7b} \\ \scriptsize{32 layers} \\ \scriptsize{4096 dimensions}}} & CLINC & 97.13$\pm$\scriptsize{0.11} & 97.60$\pm$\scriptsize{0.11} \\
 & Banking & 92.53$\pm$\scriptsize{0.12} & 92.50$\pm$\scriptsize{0.14} \\
 & DBpedia & 96.00$\pm$\scriptsize{0.08} & 95.70$\pm$\scriptsize{0.07} \\
 & HWU & 90.02$\pm$\scriptsize{0.09} & 90.43$\pm$\scriptsize{0.11} \\
 \hline
\end{tabular}
}
\caption{Comparison of final accuracy (\%) between KLDA-E and Joint Fine-tuning on text classification datasets using various LFMs.}
\label{tab:text}
\end{center}
\vspace{-4mm}
\end{table}

\begin{figure*}[t]
    \begin{center}
    \includegraphics[width=0.97\textwidth]{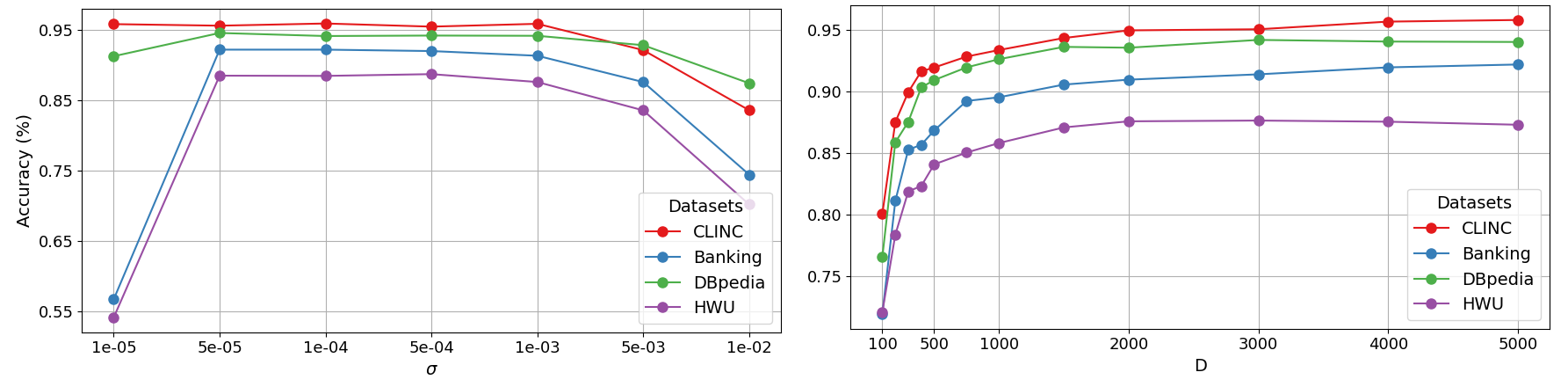}
    \caption{Hyperparameter impact on KLDA: (Left) Effect of \( \sigma \) with \( D = 5000 \). (Right) Effect of \( D \) with \( \sigma = 10^{-3} \). The FM is BART-base with 768 hidden dimensions.}
    \label{fig:KLDA_parameters}
    \end{center}
    \vspace{-2mm}
\end{figure*}

\begin{table}[h]
\begin{center}
% \resizebox{\columnwidth}{!}{
\scalebox{0.9}{ % fontsize 9
\begin{tabular}{>{\centering\arraybackslash}p{2.4cm}|>{\centering\arraybackslash}p{1.9cm}|>{\centering\arraybackslash}p{1.5cm}|>{\centering\arraybackslash}p{1.5cm}}
\hline
\textbf{Model} & \textbf{Dataset} & \textbf{KLDA} & \textbf{Joint} \\
\hline
\multirow{4}{*}{\makecell[c]{\textbf{DINOv2-small} \\ \scriptsize{12 layers} \\ \scriptsize{384 dimensions}}} & CIFAR10 & 97.00$\pm$\scriptsize{0.07} & 97.02$\pm$\scriptsize{0.09} \\
 & CIFAR100 & 84.21$\pm$\scriptsize{0.08} & 85.52$\pm$\scriptsize{0.17} \\
 & T-ImageNet & 78.67$\pm$\scriptsize{0.08} & 81.30$\pm$\scriptsize{0.17} \\
 & Cars & 81.94$\pm$\scriptsize{0.11} & 81.88$\pm$\scriptsize{0.23} \\
\hline
\multirow{4}{*}{\makecell[c]{\textbf{DINOv2-base} \\ \scriptsize{12 layers} \\ \scriptsize{768 dimensions}}} & CIFAR10 & 98.45$\pm$\scriptsize{0.04} & 98.54$\pm$\scriptsize{0.06} \\
 & CIFAR100 & 88.81$\pm$\scriptsize{0.07} & 90.30$\pm$\scriptsize{0.09} \\
 & T-ImageNet & 83.18$\pm$\scriptsize{0.11} & 86.43$\pm$\scriptsize{0.14} \\
 & Cars & 87.45$\pm$\scriptsize{0.14} & 87.47$\pm$\scriptsize{0.21} \\
\hline
\end{tabular}
}
\caption{Comparison of final accuracy (\%) between KLDA and Joint Fine-tuning on image classification datasets with different VFMs. KLDA results are without ensembles, as we didn't observe any notable improvement using ensembles on image classification datasets.}
\label{tab:image}
\end{center}
\vspace{-4mm}
\end{table}

\subsection{Efficiency and Runtime Analysis}
KLDA is highly efficient as it does not update the FM parameters or compute gradients for training. Instead, it simply computes class means and the covariance matrix. On the CLINC dataset with a BART-base LM, KLDA trains in approximately 10 seconds on our GPU setup -- comparable to the time required to extract latent features from the FM. In contrast, Joint Fine-tuning takes about 4 minutes to train. CIL fine-tuning baselines take much longer to train, as they involve updating the model incrementally on each task. They often require additional computations too, e.g., computing the output of their previous versions (KD) or generating pseudo-replay data (LAMOL and VAG), leading to training times ranging from 11 to 23 minutes. 

\subsection{Generalizability Across Different LMs}
We conducted experiments on text classification datasets with five other LMs of varying sizes. The results, shown in Table \ref{tab:text}, indicate that KLDA consistently achieves performance comparable to Joint Fine-tuning across all datasets, regardless of the LM used. This highlights the robustness of our approach for CIL.

\subsection{Evaluation on Image Datasets}
We now evaluate KLDA on image classification datasets, comparing its performance against the Joint upper bound. We do not use other baselines here as their accuracy values are significantly below the upper bound \cite{lin2023class}. The results are presented in Table \ref{tab:image}. KLDA achieves performance on par with the Joint on two datasets and performs competitively on the other two (CIFAR100 and Tiny ImageNet) with only a slight gap. This gap suggests that the current VFMs, unlike their language counterparts, still lack sufficiently expressive and generalized features for more complex datasets.

\subsection{Analysis of Hyperparameters}
KLDA's performance depends on two key hyperparameters: transform dimension \( D \) and kernel scale \( \sigma \). Figure \ref{fig:KLDA_parameters} shows how these hyperparameters affect accuracy across different text classification datasets. \( D \) controls the balance between the memory usage and the accuracy of kernel approximation. We found that setting $D$ to 5000 provides a good balance, offering sufficient accuracy without excessive memory usage. \( \sigma \) affects the scale of the RBF kernel and thus influences the separation of the transformed features. We fixed \( \sigma \) for all the datasets after determining it for each FM. KLDA performs well across all datasets with this parameter setting. This indicates that KLDA can learn various tasks incrementally without %the need for major 
adjustments to its configuration. 

\section{Conclusion}
\label{sec:conclusion}
Class-incremental learning is perhaps the most challenging setting of continual learning as it faces two major CF and ICS problems. 
KLDA avoids both challenges by utilizing a fixed foundation model and relying solely on its latent features while enhancing them with the \textit{Radial Basis Function} kernel. Since computing the full kernel matrix is infeasible in CIL, we employed \textit{Random Fourier Features}, an approximation method that enables kernel transformation while allowing incremental updates to our classification model. This is achieved by maintaining class means and a shared covariance matrix, allowing for better class separation through \textit{Linear Discriminant Analysis}. Our experiments show that KLDA significantly outperforms existing baselines and, more importantly, achieves accuracy levels on par with the upper bound accuracy of joint training. % on all classes/tasks together simultaneously.

\vspace{2mm}
\noindent
\textbf{Limitations:}
KLDA assumes that the foundation model contains sufficient features for CIL tasks in the target domain. If the FM features are not well-suited to a specific domain, the accuracy of our method may suffer. A standard approach to address this is to fine-tune or adapt the general-purpose FM using a large domain-specific corpus before applying it to CIL.

\section*{Acknowledgments}
This work was supported in part by four NSF grants (IIS-2229876, IIS-1910424, IIS-1838770, and CNS-2225427). % and a research contract from KDDI Research.

\bibliography{aaai25}

\begin{thebibliography}{57}
\providecommand{\natexlab}[1]{#1}

\bibitem[{Abati et~al.(2020)Abati, Tomczak, Blankevoort, Calderara, Cucchiara, and Bejnordi}]{abati2020conditional}
Abati, D.; Tomczak, J.; Blankevoort, T.; Calderara, S.; Cucchiara, R.; and Bejnordi, B.~E. 2020.
\newblock Conditional channel gated networks for task-aware continual learning.
\newblock In \emph{Proceedings of the IEEE/CVF Conference on Computer Vision and Pattern Recognition}, 3931--3940.

\bibitem[{Aljundi et~al.(2019)Aljundi, Lin, Goujaud, and Bengio}]{aljundi2019gradient}
Aljundi, R.; Lin, M.; Goujaud, B.; and Bengio, Y. 2019.
\newblock Gradient based sample selection for online continual learning.
\newblock \emph{Advances in neural information processing systems}, 32.

\bibitem[{Auer et~al.(2007)Auer, Bizer, Kobilarov, Lehmann, Cyganiak, and Ives}]{auer2007dbpedia}
Auer, S.; Bizer, C.; Kobilarov, G.; Lehmann, J.; Cyganiak, R.; and Ives, Z. 2007.
\newblock Dbpedia: A nucleus for a web of open data.
\newblock In \emph{international semantic web conference}, 722--735.

\bibitem[{Casanueva et~al.(2020)Casanueva, Tem{\v{c}}inas, Gerz, Henderson, and Vuli{\'c}}]{casanueva2020efficient}
Casanueva, I.; Tem{\v{c}}inas, T.; Gerz, D.; Henderson, M.; and Vuli{\'c}, I. 2020.
\newblock Efficient Intent Detection with Dual Sentence Encoders.
\newblock In \emph{Proceedings of the 2nd Workshop on Natural Language Processing for Conversational AI}, 38--45.

\bibitem[{Chen and Liu(2018)}]{chen2018lifelong}
Chen, Z.; and Liu, B. 2018.
\newblock Lifelong machine learning.
\newblock \emph{Synthesis Lectures on Artificial Intelligence and Machine Learning}, 12(3): 1--207.

\bibitem[{De~Lange et~al.(2021)De~Lange, Aljundi, Masana, Parisot, Jia, Leonardis, Slabaugh, and Tuytelaars}]{de2021continual}
De~Lange, M.; Aljundi, R.; Masana, M.; Parisot, S.; Jia, X.; Leonardis, A.; Slabaugh, G.; and Tuytelaars, T. 2021.
\newblock A continual learning survey: Defying forgetting in classification tasks.
\newblock \emph{IEEE transactions on pattern analysis and machine intelligence}, 44(7): 3366--3385.

\bibitem[{Deng et~al.(2009)Deng, Dong, Socher, Li, Li, and Fei-Fei}]{deng2009imagenet}
Deng, J.; Dong, W.; Socher, R.; Li, L.-J.; Li, K.; and Fei-Fei, L. 2009.
\newblock Imagenet: A large-scale hierarchical image database.
\newblock In \emph{2009 IEEE conference on computer vision and pattern recognition}, 248--255. Ieee.

\bibitem[{Dosovitskiy et~al.(2020)Dosovitskiy, Beyer, Kolesnikov, Weissenborn, Zhai, Unterthiner, Dehghani, Minderer, Heigold, Gelly et~al.}]{dosovitskiy2020image}
Dosovitskiy, A.; Beyer, L.; Kolesnikov, A.; Weissenborn, D.; Zhai, X.; Unterthiner, T.; Dehghani, M.; Minderer, M.; Heigold, G.; Gelly, S.; et~al. 2020.
\newblock An Image is Worth 16x16 Words: Transformers for Image Recognition at Scale.
\newblock In \emph{International Conference on Learning Representations}.

\bibitem[{Geng et~al.(2021)Geng, Yuan, Xu, Shen, Xu, and Yang}]{geng2021continual}
Geng, B.; Yuan, F.; Xu, Q.; Shen, Y.; Xu, R.; and Yang, M. 2021.
\newblock Continual Learning for Task-oriented Dialogue System with Iterative Network Pruning, Expanding and Masking.
\newblock In \emph{Proceedings of the 59th Annual Meeting of the Association for Computational Linguistics (ACL/IJCNLP)}.

\bibitem[{Gururangan et~al.(2022)Gururangan, Lewis, Holtzman, Smith, and Zettlemoyer}]{gururangan2022demix}
Gururangan, S.; Lewis, M.; Holtzman, A.; Smith, N.~A.; and Zettlemoyer, L. 2022.
\newblock DEMix Layers: Disentangling Domains for Modular Language Modeling.
\newblock In \emph{Proceedings of the 2022 Conference of the North American Chapter of the Association for Computational Linguistics}, 5557--5576.

\bibitem[{Hastie(2009)}]{hastie2009elements}
Hastie, T. 2009.
\newblock The elements of statistical learning: data mining, inference, and prediction.

\bibitem[{Hayes and Kanan(2020)}]{hayes2020lifelong}
Hayes, T.~L.; and Kanan, C. 2020.
\newblock Lifelong machine learning with deep streaming linear discriminant analysis.
\newblock In \emph{Proceedings of the IEEE/CVF conference on computer vision and pattern recognition workshops}, 220--221.

\bibitem[{He et~al.(2016)He, Zhang, Ren, and Sun}]{he2016deep}
He, K.; Zhang, X.; Ren, S.; and Sun, J. 2016.
\newblock Deep residual learning for image recognition.
\newblock In \emph{Proceedings of the IEEE conference on computer vision and pattern recognition}, 770--778.

\bibitem[{He and Jaeger(2018)}]{he2018overcoming}
He, X.; and Jaeger, H. 2018.
\newblock Overcoming catastrophic interference using conceptor-aided backpropagation.
\newblock In \emph{International Conference on Learning Representations}.

\bibitem[{Hinton, Vinyals, and Dean(2015)}]{hinton2015distilling}
Hinton, G.; Vinyals, O.; and Dean, J. 2015.
\newblock Distilling the knowledge in a neural network.
\newblock \emph{arXiv preprint arXiv:1503.02531}.

\bibitem[{Houlsby et~al.(2019)Houlsby, Giurgiu, Jastrzebski, Morrone, De~Laroussilhe, Gesmundo, Attariyan, and Gelly}]{houlsby2019parameter}
Houlsby, N.; Giurgiu, A.; Jastrzebski, S.; Morrone, B.; De~Laroussilhe, Q.; Gesmundo, A.; Attariyan, M.; and Gelly, S. 2019.
\newblock Parameter-efficient transfer learning for NLP.
\newblock In \emph{International Conference on Machine Learning}, 2790--2799. PMLR.

\bibitem[{Huang et~al.(2021)Huang, Zhang, Chen, Wang, and Yang}]{huang2021continual}
Huang, Y.; Zhang, Y.; Chen, J.; Wang, X.; and Yang, D. 2021.
\newblock Continual Learning for Text Classification with Information Disentanglement Based Regularization.
\newblock In \emph{Proceedings of the 2021 Conference of the North American Chapter of the Association for Computational Linguistics: Human Language Technologies}, 2736--2746.

\bibitem[{Izenman(2013)}]{izenman2013linear}
Izenman, A.~J. 2013.
\newblock Linear discriminant analysis.
\newblock In \emph{Modern multivariate statistical techniques: regression, classification, and manifold learning}, 237--280. Springer.

\bibitem[{Jiang et~al.(2023)Jiang, Sablayrolles, Mensch, Bamford, Chaplot, Casas, Bressand, Lengyel, Lample, Saulnier et~al.}]{jiang2023mistral}
Jiang, A.~Q.; Sablayrolles, A.; Mensch, A.; Bamford, C.; Chaplot, D.~S.; Casas, D. d.~l.; Bressand, F.; Lengyel, G.; Lample, G.; Saulnier, L.; et~al. 2023.
\newblock Mistral 7B.
\newblock \emph{arXiv preprint arXiv:2310.06825}.

\bibitem[{Ke and Liu(2022)}]{ke2022continualsurvey}
Ke, Z.; and Liu, B. 2022.
\newblock Continual learning of natural language processing tasks: A survey.
\newblock \emph{arXiv preprint arXiv:2211.12701}.

\bibitem[{Ke et~al.(2021)Ke, Liu, Ma, Xu, and Shu}]{ke2021achieving}
Ke, Z.; Liu, B.; Ma, N.; Xu, H.; and Shu, L. 2021.
\newblock Achieving Forgetting Prevention and Knowledge Transfer in Continual Learning.
\newblock \emph{Advances in Neural Information Processing Systems}, 34.

\bibitem[{Kenton and Toutanova(2019)}]{kenton2019bert}
Kenton, J. D. M.-W.~C.; and Toutanova, L.~K. 2019.
\newblock BERT: Pre-training of Deep Bidirectional Transformers for Language Understanding.
\newblock In \emph{Proceedings of NAACL-HLT}, 4171--4186.

\bibitem[{Kim et~al.(2022)Kim, Xiao, Konishi, Ke, and Liu}]{kim2022theoretical}
Kim, G.; Xiao, C.; Konishi, T.; Ke, Z.; and Liu, B. 2022.
\newblock A theoretical study on solving continual learning.
\newblock \emph{Advances in neural information processing systems}, 35: 5065--5079.

\bibitem[{Kim et~al.(2023)Kim, Xiao, Konishi, and Liu}]{kim2023learnability}
Kim, G.; Xiao, C.; Konishi, T.; and Liu, B. 2023.
\newblock Learnability and algorithm for continual learning.
\newblock In \emph{International Conference on Machine Learning}, 16877--16896. PMLR.

\bibitem[{Kirkpatrick et~al.(2017)Kirkpatrick, Pascanu, Rabinowitz, Veness, Desjardins, Rusu, Milan, Quan, Ramalho, Grabska-Barwinska et~al.}]{kirkpatrick2017overcoming}
Kirkpatrick, J.; Pascanu, R.; Rabinowitz, N.; Veness, J.; Desjardins, G.; Rusu, A.~A.; Milan, K.; Quan, J.; Ramalho, T.; Grabska-Barwinska, A.; et~al. 2017.
\newblock Overcoming catastrophic forgetting in neural networks.
\newblock \emph{Proceedings of the national academy of sciences}, 114(13): 3521--3526.

\bibitem[{Krizhevsky, Hinton et~al.(2009)}]{krizhevsky2009learning}
Krizhevsky, A.; Hinton, G.; et~al. 2009.
\newblock Learning multiple layers of features from tiny images.

\bibitem[{Larson et~al.(2019)Larson, Mahendran, Peper, Clarke, Lee, Hill, Kummerfeld, Leach, Laurenzano, Tang et~al.}]{larson2019evaluation}
Larson, S.; Mahendran, A.; Peper, J.~J.; Clarke, C.; Lee, A.; Hill, P.; Kummerfeld, J.~K.; Leach, K.; Laurenzano, M.~A.; Tang, L.; et~al. 2019.
\newblock An Evaluation Dataset for Intent Classification and Out-of-Scope Prediction.
\newblock In \emph{Proceedings of the 2019 Conference on Empirical Methods in Natural Language Processing (EMNLP-IJCNLP)}.

\bibitem[{Le and Yang(2015)}]{le2015tiny}
Le, Y.; and Yang, X. 2015.
\newblock Tiny imagenet visual recognition challenge.
\newblock \emph{CS 231N}, 7: 7.

\bibitem[{Lewis et~al.(2019)Lewis, Liu, Goyal, Ghazvininejad, Mohamed, Levy, Stoyanov, and Zettlemoyer}]{lewis2019bart}
Lewis, M.; Liu, Y.; Goyal, N.; Ghazvininejad, M.; Mohamed, A.; Levy, O.; Stoyanov, V.; and Zettlemoyer, L. 2019.
\newblock Bart: Denoising sequence-to-sequence pre-training for natural language generation, translation, and comprehension.
\newblock \emph{arXiv preprint arXiv:1910.13461}.

\bibitem[{Lin et~al.(2024)Lin, Shao, Qian, Pan, Guo, and Liu}]{lin2023class}
Lin, H.; Shao, Y.; Qian, W.; Pan, N.; Guo, Y.; and Liu, B. 2024.
\newblock Class incremental learning via likelihood ratio based task prediction.
\newblock \emph{ICML-2024}.

\bibitem[{Lin et~al.(2022)Lin, Yang, Fan, and Zhang}]{lin2022beyond}
Lin, S.; Yang, L.; Fan, D.; and Zhang, J. 2022.
\newblock Beyond not-forgetting: Continual learning with backward knowledge transfer.
\newblock \emph{Advances in Neural Information Processing Systems}, 35: 16165--16177.

\bibitem[{Liu et~al.(2021{\natexlab{a}})Liu, Yu, He, Liu, and Zhao}]{liu2021lifelong}
Liu, Q.; Yu, X.; He, S.; Liu, K.; and Zhao, J. 2021{\natexlab{a}}.
\newblock Lifelong intent detection via multi-strategy rebalancing.
\newblock \emph{arXiv preprint arXiv:2108.04445}.

\bibitem[{Liu et~al.(2021{\natexlab{b}})Liu, Eshghi, Swietojanski, and Rieser}]{liu2021benchmarking}
Liu, X.; Eshghi, A.; Swietojanski, P.; and Rieser, V. 2021{\natexlab{b}}.
\newblock Benchmarking natural language understanding services for building conversational agents.
\newblock In \emph{Increasing Naturalness and Flexibility in Spoken Dialogue Interaction: 10th International Workshop on Spoken Dialogue Systems}, 165--183.

\bibitem[{Liu et~al.(2019)Liu, Ott, Goyal, Du, Joshi, Chen, Levy, Lewis, Zettlemoyer, and Stoyanov}]{liu2019roberta}
Liu, Y.; Ott, M.; Goyal, N.; Du, J.; Joshi, M.; Chen, D.; Levy, O.; Lewis, M.; Zettlemoyer, L.; and Stoyanov, V. 2019.
\newblock Roberta: A robustly optimized bert pretraining approach.
\newblock \emph{arXiv preprint arXiv:1907.11692}.

\bibitem[{McCloskey and Cohen(1989)}]{McCloskey1989}
McCloskey, M.; and Cohen, N.~J. 1989.
\newblock {Catastrophic interference in connectionist networks: The sequential learning problem}.
\newblock In \emph{Psychology of learning and motivation}, volume~24, 109--165. Elsevier.

\bibitem[{McDonnell et~al.(2023)McDonnell, Gong, Parvaneh, Abbasnejad, and van~den Hengel}]{mcdonnell2024ranpac}
McDonnell, M.~D.; Gong, D.; Parvaneh, A.; Abbasnejad, E.; and van~den Hengel, A. 2023.
\newblock Ranpac: Random projections and pre-trained models for continual learning.
\newblock \emph{Advances in Neural Information Processing Systems (NeurIP-2023)}, 36.

\bibitem[{Oquab et~al.(2023)Oquab, Darcet, Moutakanni, Vo, Szafraniec, Khalidov, Fernandez, HAZIZA, Massa, El-Nouby et~al.}]{oquabdinov2}
Oquab, M.; Darcet, T.; Moutakanni, T.; Vo, H.~V.; Szafraniec, M.; Khalidov, V.; Fernandez, P.; HAZIZA, D.; Massa, F.; El-Nouby, A.; et~al. 2023.
\newblock DINOv2: Learning Robust Visual Features without Supervision.
\newblock \emph{Transactions on Machine Learning Research}.

\bibitem[{Qin et~al.(2022)Qin, Zhang, Lin, Liu, Li, Sun, and Zhou}]{qin2022elle}
Qin, Y.; Zhang, J.; Lin, Y.; Liu, Z.; Li, P.; Sun, M.; and Zhou, J. 2022.
\newblock ELLE: Efficient Lifelong Pre-training for Emerging Data.
\newblock In \emph{Findings of the Association for Computational Linguistics: ACL 2022}, 2789--2810.

\bibitem[{Raffel et~al.(2020)Raffel, Shazeer, Roberts, Lee, Narang, Matena, Zhou, Li, and Liu}]{raffel2020exploring}
Raffel, C.; Shazeer, N.; Roberts, A.; Lee, K.; Narang, S.; Matena, M.; Zhou, Y.; Li, W.; and Liu, P.~J. 2020.
\newblock Exploring the limits of transfer learning with a unified text-to-text transformer.
\newblock \emph{Journal of machine learning research}, 21(140).

\bibitem[{Rahimi and Recht(2007)}]{rahimi2007random}
Rahimi, A.; and Recht, B. 2007.
\newblock Random features for large-scale kernel machines.
\newblock \emph{Advances in neural information processing systems}, 20.

\bibitem[{Reimers and Gurevych(2019)}]{reimers2019sentence}
Reimers, N.; and Gurevych, I. 2019.
\newblock Sentence-BERT: Sentence Embeddings using Siamese BERT-Networks.
\newblock In \emph{Proceedings of the 2019 Conference on Empirical Methods in Natural Language Processing (EMNLP-IJCNLP)}.

\bibitem[{Rudin(2017)}]{rudin2017fourier}
Rudin, W. 2017.
\newblock \emph{Fourier analysis on groups}.
\newblock Courier Dover Publications.

\bibitem[{Serra et~al.(2018)Serra, Suris, Miron, and Karatzoglou}]{serra2018overcoming}
Serra, J.; Suris, D.; Miron, M.; and Karatzoglou, A. 2018.
\newblock Overcoming catastrophic forgetting with hard attention to the task.
\newblock In \emph{International Conference on Machine Learning}, 4548--4557. PMLR.

\bibitem[{Shao et~al.(2023)Shao, Guo, Zhao, and Liu}]{shao2023class}
Shao, Y.; Guo, Y.; Zhao, D.; and Liu, B. 2023.
\newblock Class-Incremental Learning based on Label Generation.
\newblock In \emph{Proceedings of the 61st Annual Meeting of the Association for Computational Linguistics (Short Papers)}, 1263--1276.

\bibitem[{Shin et~al.(2017)Shin, Lee, Kim, and Kim}]{shin2017continual}
Shin, H.; Lee, J.~K.; Kim, J.; and Kim, J. 2017.
\newblock Continual learning with deep generative replay.
\newblock \emph{Advances in neural information processing systems}, 30.

\bibitem[{Sun, Ho, and Lee(2019)}]{sun2019lamol}
Sun, F.-K.; Ho, C.-H.; and Lee, H.-Y. 2019.
\newblock LAMOL: LAnguage MOdeling for Lifelong Language Learning.
\newblock In \emph{International Conference on Learning Representations}.

\bibitem[{Van~de Ven and Tolias(2019)}]{van2019three}
Van~de Ven, G.~M.; and Tolias, A.~S. 2019.
\newblock Three scenarios for continual learning.
\newblock \emph{arXiv preprint arXiv:1904.07734}.

\bibitem[{Wang et~al.(2022{\natexlab{a}})Wang, Zhou, Liu, Ye, Bian, Zhan, and Zhao}]{wang2022beef}
Wang, F.-Y.; Zhou, D.-W.; Liu, L.; Ye, H.-J.; Bian, Y.; Zhan, D.-C.; and Zhao, P. 2022{\natexlab{a}}.
\newblock BEEF: Bi-compatible class-incremental learning via energy-based expansion and fusion.
\newblock In \emph{The Eleventh International Conference on Learning Representations}.

\bibitem[{Wang et~al.(2024)Wang, Zhang, Su, and Zhu}]{wang2024comprehensive}
Wang, L.; Zhang, X.; Su, H.; and Zhu, J. 2024.
\newblock A comprehensive survey of continual learning: theory, method and application.
\newblock \emph{IEEE Transactions on Pattern Analysis and Machine Intelligence}.

\bibitem[{Wang et~al.(2023)Wang, Liu, Ji, Wang, Wu, Jiang, Chao, Han, Wang, Shao et~al.}]{wang2023rehearsal}
Wang, Z.; Liu, Y.; Ji, T.; Wang, X.; Wu, Y.; Jiang, C.; Chao, Y.; Han, Z.; Wang, L.; Shao, X.; et~al. 2023.
\newblock Rehearsal-free continual language learning via efficient parameter isolation.
\newblock In \emph{Proceedings of the 61st Annual Meeting of the Association for Computational Linguistics}, 10933--10946.

\bibitem[{Wang et~al.(2022{\natexlab{b}})Wang, Zhang, Lee, Zhang, Sun, Ren, Su, Perot, Dy, and Pfister}]{wang2022learning}
Wang, Z.; Zhang, Z.; Lee, C.-Y.; Zhang, H.; Sun, R.; Ren, X.; Su, G.; Perot, V.; Dy, J.; and Pfister, T. 2022{\natexlab{b}}.
\newblock Learning to prompt for continual learning.
\newblock In \emph{Proceedings of the IEEE/CVF Conference on Computer Vision and Pattern Recognition}, 139--149.

\bibitem[{Wortsman et~al.(2020)Wortsman, Ramanujan, Liu, Kembhavi, Rastegari, Yosinski, and Farhadi}]{wortsman2020supermasks}
Wortsman, M.; Ramanujan, V.; Liu, R.; Kembhavi, A.; Rastegari, M.; Yosinski, J.; and Farhadi, A. 2020.
\newblock Supermasks in superposition.
\newblock \emph{Advances in Neural Information Processing Systems}, 33: 15173--15184.

\bibitem[{Yan, Xie, and He(2021)}]{yan2021dynamically}
Yan, S.; Xie, J.; and He, X. 2021.
\newblock Der: Dynamically expandable representation for class incremental learning.
\newblock In \emph{Proceedings of the IEEE/CVF Conference on Computer Vision and Pattern Recognition}, 3014--3023.

\bibitem[{Yang et~al.(2015)Yang, Luo, Change~Loy, and Tang}]{yang2015large}
Yang, L.; Luo, P.; Change~Loy, C.; and Tang, X. 2015.
\newblock A large-scale car dataset for fine-grained categorization and verification.
\newblock In \emph{Proceedings of the IEEE conference on computer vision and pattern recognition}, 3973--3981.

\bibitem[{Yang et~al.(2024)Yang, Zhou, Ding, Huai, Liu, Chen, Xie, and He}]{yang2024recent}
Yang, Y.; Zhou, J.; Ding, X.; Huai, T.; Liu, S.; Chen, Q.; Xie, Y.; and He, L. 2024.
\newblock Recent advances of foundation language models-based continual learning: a survey.
\newblock \emph{ACM Computing Surveys}.

\bibitem[{Zenke, Poole, and Ganguli(2017)}]{Zenke2017continual}
Zenke, F.; Poole, B.; and Ganguli, S. 2017.
\newblock Continual learning through synaptic intelligence.
\newblock In \emph{International Conference on Machine Learning}, 3987--3995. PMLR.

\bibitem[{Zhou et~al.(2024)Zhou, Sun, Ning, Ye, and Zhan}]{zhou2024continual}
Zhou, D.-W.; Sun, H.-L.; Ning, J.; Ye, H.-J.; and Zhan, D.-C. 2024.
\newblock Continual learning with pre-trained models: A survey.
\newblock \emph{arXiv preprint arXiv:2401.16386}.

\end{thebibliography}
\end{document}